\documentclass[10pt,twocolumn,letterpaper]{article}

\usepackage{cvpr}
\usepackage{times}
\usepackage{epsfig}
\usepackage{graphicx}
\usepackage{amsmath}
\usepackage{amssymb}
\usepackage{array}
\usepackage{mathrsfs}
\usepackage[table]{xcolor}
\usepackage{booktabs}
\usepackage{setspace}
\definecolor{mygray}{gray}{.92}

\usepackage[pagebackref=false,breaklinks=true,letterpaper=true,citecolor=black,colorlinks,linkcolor=black,bookmarks=false]{hyperref}

\cvprfinalcopy 

\def\cvprPaperID{256} 

\begin{document}

\title{Semantic Correlation Promoted Shape-Variant Context for Segmentation}

\author{
Henghui Ding$^1$
\qquad
Xudong Jiang$^1$
\qquad
Bing Shuai$^2$
\qquad
Ai Qun Liu$^1$
\qquad
Gang Wang$^3$\\
\vspace{-0.45em}
\\
$^1$School of Electrical and Electronic Engineering, Nanyang Technological University, Singapore\\
$^2$Amazon, Seattle, United States \qquad $^3$Alibaba Group, Hangzhou, China\\
{\tt\small \{ding0093, exdjiang, eaqliu\}@ntu.edu.sg, \{beinshuai, gangwang6\}@gmail.com}
}

\maketitle

\begin{abstract}
   Context is essential for semantic segmentation. Due to the diverse shapes of objects and their complex layout in various scene images, the spatial scales and shapes of contexts for different objects have very large variation. It is thus ineffective or inefficient to aggregate various context information from a predefined fixed region. In this work, we propose to generate a scale- and shape-variant semantic mask for each pixel to confine its contextual region. To this end, we first propose a novel paired convolution to infer the semantic correlation of the pair and based on that to generate a shape mask. Using the inferred spatial scope of the contextual region, we propose a shape-variant convolution, of which the receptive field is controlled by the shape mask that varies with the appearance of input. In this way, the proposed network aggregates the context information of a pixel from its semantic-correlated region instead of a predefined fixed region. Furthermore, this work also proposes a labeling denoising model to reduce wrong predictions caused by the noisy low-level features. Without bells and whistles, the proposed segmentation network achieves new state-of-the-arts consistently on the six public segmentation datasets.
\end{abstract}

\section{Introduction}

Semantic segmentation or scene parsing is aimed at classifying every pixel in scene images to one of the predefined categories (e.g., person, car, etc.). It has been a critical element in artificial intelligence and can be applied in many practice applications, such as automatic parking system. The recent success of Deep Neural Networks has greatly improved the performance of semantic segmentation~\cite{chen2016deeplab, Lin_2017_CVPR, Zhao_2017_CVPR, chen2018encoder}. Most of state-of-the-arts segmentation networks are based on Convolutional Neural Networks (CNNs)~\cite{krizhevsky2012imagenet, simonyan2014very, szegedy2015going, he2016deep, Huang_2017_CVPR} pre-trained on ImageNet~\cite{russakovsky2015imagenet}, in which CNNs are employed as the local feature extractor. To achieve robust semantic segmentation, informative high-level context is necessary. Context provides surrounding information of the object, which helps a better discrimination of the object.

\begin{figure}[t]
\centering
  \includegraphics[width=0.476\textwidth]{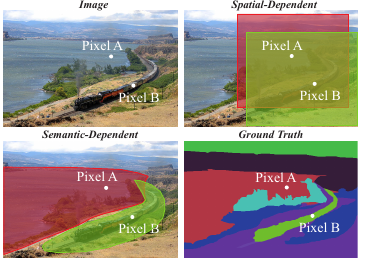}
\caption{Most existing methods model spatial-dependent context with predefined window (e.g., the red rectangle region for pixel A in the second image), which may not provide the specific context information and thus weaken the semantic shape layout. In this work, we propose to aggregate the context information from the semantic-dependent region instead of the spatial-dependent one.}
\label{Figure:introduction}
\end{figure}

However, due to the diverse shapes (including contours, scales, etc) and the complex layout of objects in scene images, the common context aggregated within a predefined fixed region weakens the semantic shape layout and might bring unnecessary irrelevant information. For example in Figure \ref{Figure:introduction}, contexts of pixel A (lake) and pixel B (train) should be different but the predefined receptive fields to collect their contexts largely overlap, which decrease their discriminative power. Meanwhile, not all information in a predefined surrounding region (rectangle region in the second image) is beneficial for their final parsing. Information collected in the semantic correlated region is more helpful to identify the object while that in an irrelevant region, though spatially close, may result in parsing error and hence should be suppressed or even disregarded. For pixel A in Figure \ref{Figure:introduction}, information of pixels belonging to the lake and its shore, which are semantic correlated, might be more beneficial than information of other pixels in the predefined fixed window. And for pixel B, the desirable shape of context would align with the train and railroad track. Besides, uniformly integrating smooth global information would degrade the location identity and the local discriminative features~\cite{yan2016combining}. Therefore, for better scene parsing, diverse shapes of semantic consistent context should be customized. Most existing methods tend to model statistical average representation among a fixed rectangle region~\cite{chen2016deeplab, Ding_2018_CVPR, yu2015multi, yang2018denseaspp} or the global region~\cite{Yu_2018_CVPR, Zhang_2018_CVPR, hung2017scene}. In this work, by taking into account the semantic correlation and the shape layout of objects, we propose a shape-variant context model to aggregate the surrounding information of each pixel from their semantic-correlated region inferred according to the appearance of input image.

To this end, we propose first to learn the relation between a target pixel and others by a novel paired convolution followed by a Gaussian mapping function. The learnt network produces higher value for two pixels with stronger semantic correlation and lower value for weaker correlation. Thus, the proposed network will generate a shape mask indicating a semantic correlated region for each pixel. With the shape mask specifying the size and shape of desirable receptive field, we further propose a shape-variant convolution to aggregate context from semantic-dependent region. The shape-variant convolution is specified by a set of learnable location-invariant convolution parameters and the location-variant shape masks. Thus, the parameters are dependent variables of the semantic correlated region of input image, which change with varying shapes and scales of objects.
Furthermore, since the shape-variant context is implicitly scale-variant, we can model not only multi-shape but also multi-scale information in a single layer instead of paralleled~\cite{chen2016deeplab, Zhao_2017_CVPR, yang2018denseaspp} or stacked~\cite{yu2015multi, Ding_2018_CVPR} multi-layers. From a macro perspective, the proposed approach helps control the information flow within network through learning the semantic and spatial relationship of features and determine the information passing or suppression.

The proposed scale- and shape-variant context model enhances the discriminative power of the high level features. Higher level features are more robust to noise than lower level features at a price of lower spatial position sensitivity. Thus, many segmentation networks also aggregate low-level features to improve the position accuracy of the segmentation~\cite{long2015fully, chen2018encoder, Ding_2018_CVPR, hariharan2015hypercolumns}. However, aggregating low-level features, though helps to recover spatial information, may bring some debatable noise sensitive information that leads to a wrong classification of some pixels. Thus, we propose a model that utilizes the higher level features to attenuate the noisy information of the low-level features before aggregating them, i.e. signals denoising. In such a way, the network could better exploit the advantages of low-level features by alleviating their problems.

In summary, this paper makes the following contributions: 1) we propose a novel paired convolution to infer the semantic correlation of two pixels and based on that to generate a semantic-correlated region for each pixel; 2) we propose a shape-variant context aggregated within the semantic-correlated region to model the diverse shapes and scales of contexts, which greatly enhances the modeling ability of network; 3) we propose a labeling denoising model to reduce the labeling errors caused by the noisy low-level features; 4) we achieve new state-of-the-art performance consistently on six public semantic segmentation datasets.

\section{Related work}
\label{Section:ReviewFCN}

Recently, Deep Neural Networks has achieved great success on computer vision~\cite{he2016deep, gu2018recent, gu2018unpaired, liu2019feature, liu2018mcn, gu2019scene, liu2019cse, tay2019aabet}. Based on the Fully Convolutional Network (FCN)~\cite{long2015fully}, in which the fully connected layers in original CNNs are converted to convolutional layers, a lot of approaches, e.g.,~\cite{chen2016deeplab, ghiasi2016laplacian, noh2015learning, islam2017gated, liu2015semantic, Zhao_2018_ECCV, bulo2017loss, shuai2019toward}, are proposed to improve the performance of semantic segmentation.



\textbf{Contextual features modeling} plays an important role in scene parsing. \cite{lucchi2011spatial} shows that global spatial information helps enforce the features consistency. DeepLab~\cite{chen2016deeplab} proposes atrous spatial pyramid pooling (ASPP) to aggregate multi-scale image representations from parallel branches with different dilated rate. DilatedNet~\cite{yu2015multi} appends several dilated convolution layers after the score maps to perform multi-scale context aggregation. DAG-RNN~\cite{shuai2017scene} and Byeon~\cite{Byeon_2015_CVPR} propose to model long-range context through recurrent neural networks. Zoom-out~\cite{mostajabi2015feedforward} proposes a feed-forward architecture to extract hierarchical zoom-out features. CRF-RNN~\cite{zheng2015conditional} uses recurrent layers  for jointly end-to-end training the dense CRFs~\cite{krahenbuhl2011efficient} with their segmentation networks. Piecewise~\cite{lin2016efficient} formulates CNN-based pairwise potential functions to capture patch-patch context and designs image pyramid input for patch-background context. PSPNet~\cite{Zhao_2017_CVPR} introduces pyramid spatial pooling (PSP) to perform different-region-based global information aggregation. Recently, CCL~\cite{Ding_2018_CVPR} proposes a context contrasted local model to parallelly collect local and its surrounding information. EncNet~\cite{Zhang_2018_CVPR} encodes semantic context to network and stress class-dependent feature maps.

Different from previous methods, in this work, we try to aggregate context information from semantic-closer region but suppress irrelevant information even in the spatial-closer region. We propose a shape-adaptive convolutional layer to learn diverse shaped contexts whose shapes are determined by the object shape, scale and its surrounding support of the input image. The proposed approach is aimed at not only retaining the location identity and layout information but also building the effective semantic correlation shown in the training images.

\textbf{Label variety}  is another challenging problem in semantic segmentation. PSPNet~\cite{Zhao_2017_CVPR} observes the confusion categories and demonstrates that the PSPNet can better address the confusion labels than FCN\cite{long2015fully}. Geng et al.~\cite{geng2018network} propose to infer the discriminative confusing groups from a prior confusion matrix. DFN~\cite{Yu_2018_CVPR} introduces a smooth and border network to tackle confusing classes. Davis et al.~\cite{davis2018classification} propose to refine the parsing results using Bayesian strategy with confusion probabilities and label priors. Huang et al.~\cite{huang2017error} propose a LabelReplacement network to correct the error predictions. Different from these methods, we propose a learnable Labeling Denoising (LD) model to solve the problems of confusion labels by utilizing the robust higher level features to attenuate the noise in lower level features.
\section{The Proposed Approach}

\subsection{Semantic Correlation Dependent Context}

Semantic Segmentation needs to simultaneously deal with object recognition and localization, and hence should build the dense feature connections among large region as well as retain the location identity. Meanwhile, due to the diverse shapes and complicated layout of objects in scene images, the scales and shapes of contexts for different objects are supposed to have very large variation.

\begin{figure}[t]
\centering
  \includegraphics[width=0.476\textwidth]{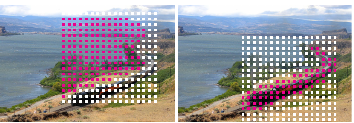}
\caption{(Best viewed in color) We propose a novel semantic correlation dependent shape-variant context, which boosts the semantic-correlated features (magenta color) while suppresses the others (white color).}
\label{Figure:Shape_Gates}
\vspace{-0.3cm}
\end{figure}

Many existing context modeling methods tend to aggregate surrounding information with a fixed size of rectangle window across all locations, which weaken the location identity and might not be able to effectively represent the diverse shapes and scales of objects in scene images.
Different from previous works, we put forward that the more desirable context region should be shape-variant according to the shape of the object and its background that support the object. For example, for pixels belong to the train in Figure \ref{Figure:Shape_Gates}, the more beneficial context should be the surrounding information along the railroad track (magenta color), which are closer in semantics than in space. In a word, for different pixel positions, surrounding information should be collected from semantic correlated region that supports the existence of the correct class of this pixel. Therefore, in this work, we propose a semantic correlation dependent shape-variant context (SVC) to model diverse shape/scale contexts with location identity. In the SVC, context aggregation is controlled by a semantic correlation mask, specifying where the information should be collected with what extent. With the semantic mask, features in semantic correlated regions are boosted and other irrelevant ones are suppressed. Thus, better context information for parsing of each pixel is aggregated within the specific shape region that supports the existence of the correct class of the object.

\vspace{0.76em}
\noindent \textbf{Learn the Semantic Correlation}
\vspace{0.12em}

\begin{figure}[t]
\centering
  \includegraphics[width=0.476\textwidth]{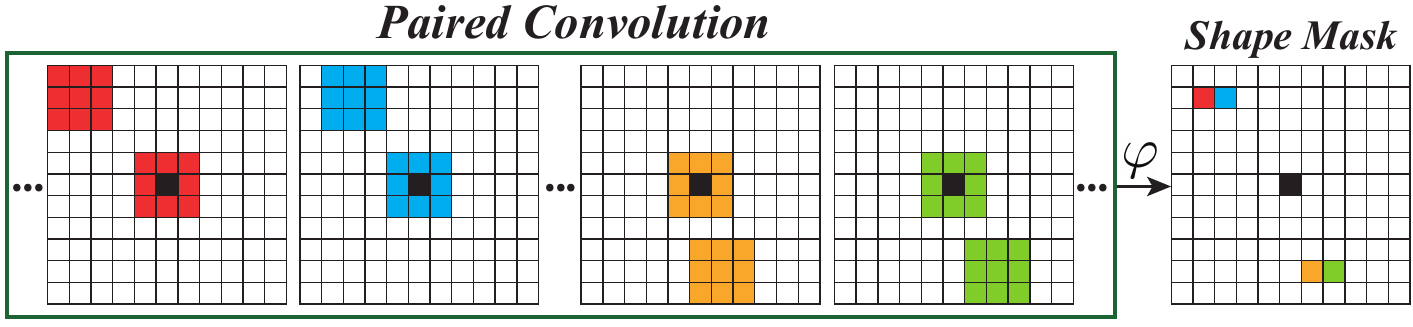}
\caption{(Best viewed in color) A shape mask is inferred by the Paired Convolution and Gaussian mapping function $\varphi$, which are designed to learn the semantic correlation between the target pixel and other pixels within the shape mask. Here we show an example of 4 values in a 11$\times$11 mask of the target pixel (dark), in which 4 mask values are generated by the 4 filters of the same color.}
\label{Figure:Paired_Conv}
\vspace{-0.13cm}
\end{figure}

The shape masks that represent semantic correlation trim the context shapes and decide where the information should be collected with what extent. Next we discuss how to learn the semantic correlation, i.e. how to generate the semantic shape masks. Each value in a shape mask represents the correlation of the corresponding pixel to the target pixel (the center pixel of the mask). Thus, the semantic relationship of each pixel with the target pixel need be learned and injected to the corresponding position in the shape mask. To this end, we introduce a paired convolution, as shown in Figure \ref{Figure:Paired_Conv}, in which a pair of $3$$\times$$3$ local convolution with specific relative position are employed to learn the semantic and spatial correlation of the corresponding pixel with the target pixel (the central dark pixel in Figure \ref{Figure:Paired_Conv}). In each filter of the paired convolution, there are a central convolution for the target pixel and another convolution whose position corresponds to the position in shape mask for the corresponding pixel. We have observed and hence assumed that the feature appearances of pixels belonging to the same object and its context will show strong correlation because they frequently coexist in the training images. Therefore, the difference of the two convolution outputs $\mathcal{D}_{m,n}^{i,j}$ can be minimized for two pixels belonging to the same object and its context by learning the convolution parameters from the training images.

\begin{equation}\label{paired_conv}
\mathcal{D}_{m,n}^{i,j}={\mathcal{F}}^{i,j}\ast\Theta_{0,0}^{m,n}-{\mathcal{F}}^{i-m,j-n}\ast\Theta_{{m,n}}^{m,n}
\end{equation}
where $\ast$ donates the local convolution operator, $\mathcal{D}_{m,n}^{i,j}$ represents the convolution output discrepancy of $(i-m, j-n)$ to target position $(i, j)$, ${\mathcal{F}}^{i,j}$ and ${\mathcal{F}}^{i-m,j-n}$ are local features at position $(i,j)$ and $(i-m,j-m) $, $\Theta_{0,0}^{m,n}$ and $\Theta_{m,n}^{m,n}$ are corresponding parameters of two local kernels in the paired convolution. As there might be negative or positive difference, we map it to the value of our context shape mask by a Gaussian function:
\begin{equation}\label{gausMapping}
\mathcal{M}_{m,n}^{i,j}=\varphi(\mathcal{D}_{m,n}^{i,j})
\end{equation}
where $\varphi(a)$$=$$exp(-\frac{a^2}{\sigma^2})$, which maps the  convolution output discrepancy to a semantic correlation value. A smaller discrepancy generates a higher semantic correlation value. $\mathcal{M}_{m,n}^{i,j}$ is the mask value at position ($m$, $n$) in the semantic shape mask of pixel ($i$, $j$). Note that the result is not sensitive to $\sigma$ as the parameters of the two convolutions are learnt. We use $\sigma=3$ in our experiments.


\vspace{0.76em}
\noindent \textbf{Shape-Variant Context}
\vspace{0.12em}

The goal of shape-variant context is to customize a desirable shape/scale of context for each pixel instead of a simple smooth context information. To achieve this, we further propose a shape-variant convolution (SV Conv) to adaptively collect the surrounding information. The parameters of shape-variant convolution consists of location-invariant learnable convolution parameters and semantic shape mask inferred by the proposed paired convolution. The shape mask is used to control the receptive field of the convolution process for each position according to the semantic correlation. Such shape mask crops the convolution kernels into different shapes/scales and leads to a shape-variant convolution operation. In such a way, the proposed method greatly enhances the network modeling ability of diverse shape context.


\begin{figure}[t]
\centering
  \includegraphics[width=0.48\textwidth]{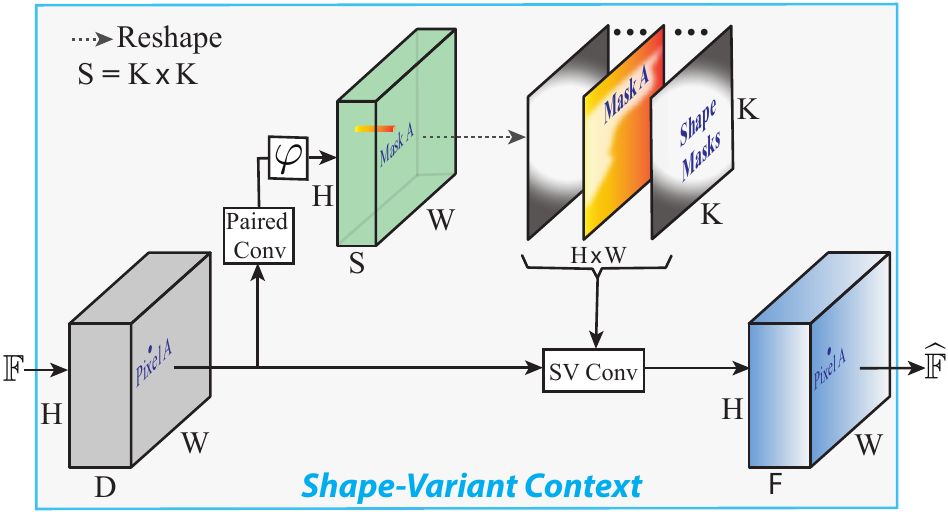}
\caption{Semantic correlation-dependent shape-variant context aggregates surrounding information according to the semantic correlation and hence customizes an effective contextual region. It helps control the information flow within network via deciding what information to be passed or suppressed.}
\label{Figure:Shape-Variant-Context}
\end{figure}

The proposed shape-variant context is shown in Figure \ref{Figure:Shape-Variant-Context}. There are two branches, the bypass is designed to learn the semantic correlation, whose outputs are then input to the shape-variant convolution (SV Conv) to provide the semantic shape mask.
In detail, the side branch employs paired convolution to learn from the local feature $\mathbb{F}$ from a pre-trained CNN the correlations of each pixel with each of all other pixels within the kernel of size \emph{$K$$\times$$K$} centered at this pixel, as described in Eq.$\!$ (\ref{paired_conv}) and Eq.$\!$ (\ref{gausMapping}).  The number of the output channels is \emph{$S$$ =$$K$$\times$$K$} where \emph{$K$$\times$$K$} is the kernel size of the proposed shape-variant convolution.
The semantic shape masks inferred form input features are employed to weight the normal learnable convolutions ($F$ filters) parameterized by $\theta_{m,n}^{d,f}$ of the main branch by:
\begin{equation}
\hat{\theta}^{i,j,d,f}_{m,n} = \mathcal{M}_{m,n}^{i,j}\theta_{m,n}^{d,f} \label{eq3}
\end{equation}
where $\theta_{m,n}^{d,f}$$\in$${\Theta}$ is the convolution parameter for $d^{th}$ input channel at position $(m,n)$ of the $f^{th}$ normal location-invariant learnable filter, $d$$\in$${(1,2...,D)}$ and $f$$\in$$ {(1,2...,F)}$. The filter kernel size is \emph{$K$$\times$$K$} and   $(i,j)$ is the index of feature map position across all the \emph{$H$$\times$$W$} positions. By Eq.$\!$ (3), the receptive field of the $F$ normal learnable convolutions parameterized by $\theta_{m,n}^{d,f}$ is transformed from a fixed size of \emph{$K$$\times$$K$} to effectively different sizes and shapes for different pixels $(i,j)$ determined by the proposed semantic shape masks $\mathcal{M}_{m,n}^{i,j}$. The resulting $F$ shape-variant filters are employed to generate variegated shape context for each spatial position $(i,j)$:
\begin{equation}
\widehat{\mathcal{F}}^{i,j,f}=\sum_{d=1}^{D}\sum_{m=-K_1}^{K_1}\sum_{n=-K_1}^{K_1}\hat{\theta}_{m,n}^{i,j,d,f}\mathcal{F}^{i-m,j-n,d}
\label{eq4}
\end{equation}
where $K_1$$=$$(K-1)/2$ and $\mathcal{F}^{i-m,j-n,d} \in \mathbb{F}$. ${\widehat{\mathcal{F}}}^{i,j ,f} \in \widehat{\mathbb{F}}$ is one target contextual feature map. In such a way, $\mathcal{M}_{m,n}^{i,j}$ instructs how to collect $\mathcal{F}^{i-m,j-n}$ for $\widehat{\mathcal{F}}^{i,j}$ during convolution.
All of these functions are differentiable and its back-propagation is easy to derive.

The standard convolutional operation is location invariant and do not vary with testing images after training. Thus, it could not customize different shapes/scales of context information for different objects of input images. The proposed SV Conv consists of a learnable location-invariant convolution and a location-variant semantic shape mask inferred from the input image. The former is to model the statistical average of the spatial-channel distribution and the latter is to determine the size and shape of the convolution receptive field. They together function as a shape-variant operator to better model the shaped context.


\vspace{0.76em}
\noindent \textbf{Modeling Diverse Shapes in a Single Layer}
\vspace{0.12em}

Due to the diverse shapes/scales of contextual regions and the shape constraint of convolutional kernels, it is difficult to use a single normal convolutional layer to effectively model shape-variant context because the scope of the context, including its scale and contour, dramatically changes for different objects of input images. With the proposed shape-variant semantic correlation masks, convolution regions with diverse shapes and scales are specified and thus we could model multi-shape and multi-scale information in a single layer.

\vspace{0.76em}
\noindent\textbf{Comparison with State-of-the-art Context Models}
\vspace{0.12em}

Different from the previous context methods that tend to model statistical average representation within a predefined rectangle region, e.g.,~\cite{chen2016deeplab, yu2015multi, Zhao_2017_CVPR, Ding_2018_CVPR, yang2018denseaspp, Peng_2017_CVPR}, the proposed approach utilizes the semantic correlation and intentionally picks up the relevant information according to the semantic shape mask inferred by the feature appearance of testing image. Thus, it could not only retain the shape and location identity but also effectively build the beneficial connection among correlated features for classification. Comparing with the deformable convolution~\cite{Dai_2017_ICCV}, the objective of the proposed approach is similar to it in terms of collecting the relevant information in the convolution. However, different from~\cite{Dai_2017_ICCV} that tries to achieve this via deforming the sampling locations, the proposed SVC finds out the semantic correlation to enhance or attenuate the corresponding information, explicitly leading to the shape- and scale-variant modeling. The criteria or the methods to find the relevant information in the two approaches are also different. Furthermore, our method models diverse shapes of semantic-dependent context in a single layer without stacking layers in~\cite{Dai_2017_ICCV}, and avoids the ``atrous'' in deformable convolution that may lose some detailed information.

\subsection{Labeling Denoising}

\begin{figure}[t]
\centering
  \scriptsize{\hspace{4.6em} \textbf{Image}  \hfill \hspace{3.5em} \textbf{Labeling Errors} \hfill \hspace{1.25em} \textbf{Ground Truth} \hfill \hspace{6em}} \\
  \includegraphics[width=0.476\textwidth]{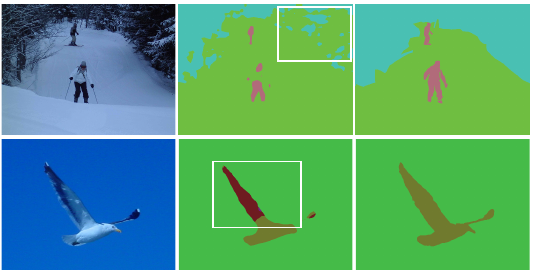}
\vspace{-0.24cm}
\caption{In-context error, e.g., the first row, refers to incorrect labeling within the label set of the image. Out-context error, e.g., the second row, refers to incorrect labeling outside the label set of the image.}
\label{Figure:LabelError}
\vspace{-0.12cm}
\end{figure}

Due to the label variety and the complicated correlation among labels in segmentation datasets, regular errors can be found in the results of most state-of-the-arts segmentation networks~\cite{chen2016deeplab, Zhao_2017_CVPR, davis2018classification, chen2018encoder, Lin_2017_CVPR}. These regular errors could be categorized into ``in-context'' error and ``out-context'' error, as show in Figure \ref{Figure:LabelError}. The ``in-context'' error is mainly caused by inaccurate positioning and inter-context influence while the ``out-context'' error is mainly caused by inaccurate classification. The proposed shape-variant context aggregates information from specific semantic correlated region, which helps mitigate the in-context and out-context labeling errors. To get elaborate spatial information, lower-level features from middle layers of CNNs are important in the encoder-decoder architecture~\cite{long2015fully, chen2018encoder, Ding_2018_CVPR, hariharan2015hypercolumns} as they contain more information about where these objects are~\cite{ghiasi2016laplacian, long2015fully}. But these low-level features also bring debatable noisy information that results in out-context errors. In contrast, the high-level features, e.g., the shape-variant context in this work, though less sensitive to the spatial location, are more robust to noise and more aware of what categories existing in a scene image. To better combine ``what'' and ``where'', we propose a labeling denoising (LD) model in this work that attenuates the noise information when extracting low-level spatial information from middle layers. 

\begin{figure}[t]
\centering
  \includegraphics[width=0.47\textwidth]{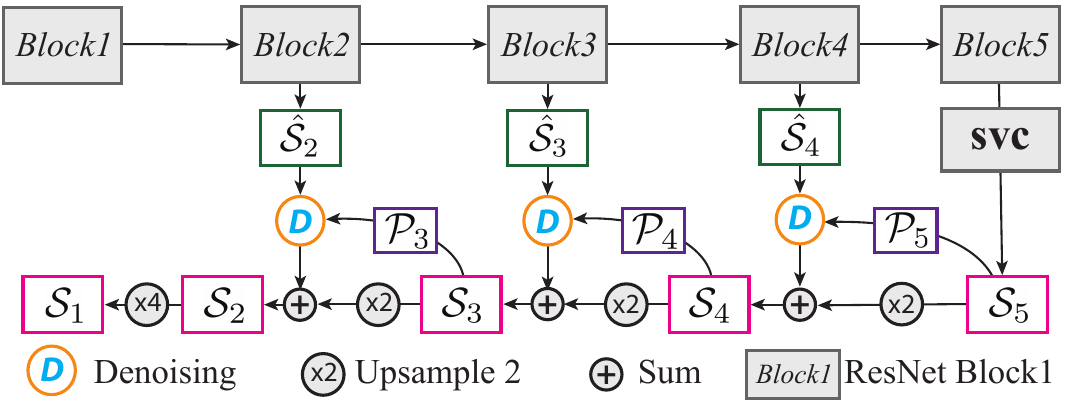}
\caption{Network Architecture. We use ResNet-101 as our base model for fine-tuning and FCN-4s as the backbone segmentation framework. LD is used in decode process for denoising.}
\vspace{-0.24cm}
\label{Figure:Network_Architecture}
\end{figure}

The labeling denoising model first infers the existence potential of each category from a higher-level block and learns penalty scores from the existence potentials. Then, the score maps generated from a lower-level block are charged by the penalty scores. Using the penalty scores learnt from the higher-level block, the scores of nonexistent categories of an input image generated by the lower-level block are greatly suppressed. First, the existence potential is inferred by the score maps from a higher-level block by:
\begin{equation}
\mathcal{E}_k =\mathit{F}_{g}(\mathit{F}_{sf}(\mathcal{S}_k))
\end{equation}
where $\mathcal{S}_k$ is the score maps from a higher-level (level $k$) block, $\mathit{F}_{sf}$ is softmax and $\mathit{F}_{g}$ is global max pooling. $\mathcal{E}_k =(e_k^1,..., e_ k^c, ..., e_ k^C)$ and $e_ k^c$ is the existent potential for class $c$ inferred by level $k$. Then, the penalty $\mathcal{P}^c$ is learnt by:
\begin{equation}
\mathcal{P }_k^c=\textrm{ReLU}(\mathcal{T}-e_ k^c)\Delta_k^c
\end{equation}
where $\mathcal{T}$  is a penalty threshold and $\Delta_k^c$ is a learnable penalty parameter. The penalty threshold and function $\textrm{ReLU}$ are used to keep the score distribution of existent classes unchanged. The penalty $\mathcal{P}_k^c$ is used to modify the score map of its next lower-level block before aggregating it into the upsampled score map of its next higher-level block as:
\begin{equation}
\mathcal{{S}}_{k-1}^c= \textrm{ReLU}(\hat{\mathcal{S}}_{k-1}^c-\mathcal{P}_k^c)+ \mathcal{S}_k^c
\end{equation}
where $\hat{\mathcal{S}}_{k-1}^c$ is the score map of class $c$ directly from the lower-level block. $\mathcal{S}_{k-1}^c$ is the denoised and aggregated score map from the highest level to the level $k-1$, which is further used to modify and be aggregated to the score maps of the lower-level blocks as Eqs (5), (6), and (7). The proposed labeling denoising (LD) model is shown in Figure \ref{Figure:Network_Architecture}. In such a way the network could take advantage of both high-level features and low-level features, i.e. better combine ``what'' and ``where''.
For the skip layers from low-level features, the scores of nonexistent categories in an input image are attenuated and those of the existent categories are retained and supplemented to score maps for positioning enhancement. As this mechanism is included in the end-to-end training process, less noisy scores are taken into account during training, and gradients for training such noisy information could be saved for other things such as positioning.
The proposed approach could also be viewed as some kind of dropout, which applies dropout to connections that reach some conditions.

\section{Experiments}
\vspace{-0.1cm}

We evaluate the proposed approach on six public benchmark, COCO-Stuff, SIFT-Flow, CamVid, PASCAL-Person-Part, PASCAL-Context, and Cityscapes.
We use ResNet-101\cite{he2016deep} pre-trained on ImageNet~\cite{russakovsky2015imagenet} as our base model for fine-tuning and FCN-4s as the backbone framework. During training, the proposed Network is trained end-to-end using standard SGD with batch size 8, fixed momentum 0.9 and weight decay 0.0005. Data augmentations like random flipping, random resize between 0.8 and 1.2 and mean subtraction are used in training. Inspired by~\cite{chen2016deeplab}, we use the "poly'' learning rate and set the initial learning rate to $5\times10^{-3}$ for newly initialized parameters and $10^{-4}$ for parameters of pretrained layers, the power is set to 0.9. Batch Normalization~\cite{ioffe2015batch} is used in new added layers to accelerate training process.
The performance is evaluated by standard pixel accuracy (pixel acc.), mean class accuracy (mean acc.) and mean Intersection-over-Union (mean IoU). Please refer to~\cite{long2015fully} for mathematical definitions.

To model diverse semantic shapes in a single layer, larger kernel is required due to the dramatic changed shapes/scales of objects. But very large kernel is resource-intensive and difficult to converge.
To address this issue, we modify Eq. (4) of the proposed SVC similar to the depthwise separable convolution~\cite{Chollet_2017_CVPR}. The simplified computation of Eq.$\!$ (4) allows us using large kernel size to model the diverse shapes in spatial space followed by a pointwise convolution to learn the cross-channel correlation. And in labeling denoising  model, we use ascending penalty thresholds $\mathcal{T} = t, 2t, 4t$, from the highest to the lowest blocks, where $t = \frac{1}{C}$ and $C$ is the number of classes.

\subsection{Ablation Study}
\label{Ablation_Study}
\begin{table}[t]
\setlength{\abovecaptionskip}{0pt}%
\setlength{\belowcaptionskip}{-10pt}
\footnotesize
\begin{center}
\begin{tabular}{l|cc}
\toprule
Methods &  PASCAL-Context &  COCO-Stuff\\
\hline
\hline
Baseline &  42.7 &  31.5\\
Baseline+SVC &  52.4 &  38.5\\
Baseline+SVC+LD &  \textbf{53.2} &  \textbf{39.6}\\
\bottomrule
\end{tabular}
\end{center}
\vspace{-0.16cm}
\caption{Ablation study of the proposed approach in terms of IoU.}
\label{Table:AblationStudy}
\end{table}

\begin{table}[t]
\footnotesize
\begin{center}
\begin{tabular}{p{2.16cm}<{\centering}|p{2.1cm}<{\centering}p{2.1cm}<{\centering}}
\toprule
Kernel Size &  SFC &  SVC\\
\hline
\hline
$0\times0$ & 42.7 & 42.7 \\
$7\times7$ & 45.6 & 48.5 \\
$11\times11$ & 46.6 & 49.4 \\
$15\times15$ & \colorbox{mygray}{47.1} & 51.2 \\
$19\times19$ & 47.0 & 52.1 \\
$23\times23$ & 46.7 & \colorbox{mygray}{\textbf{52.4}} \\
$27\times27$ & 46.5 & 52.3 \\
\bottomrule
\end{tabular}
\end{center}
\vspace{-0.12cm}
\caption{Ablation study of the proposed shape-variant context (SVC) approach by comparing it with the shape-fixed context (SFC) on different kernel size. It also shows that the performance gain is not simply brought by the increased number of parameters.}
\vspace{-0.24cm}
\label{Table:LargeKernelSize}
\end{table}

In this section we do ablation studies of the proposed shape-variant context (SVC) and labeling denoising (LD). As shown in Table \ref{Table:AblationStudy}, comparing the performance gain brought by the proposed LD on PASCAL-Conext (59 classes) and COCO-Stuff (171 classes), we can conclude that the LD could mitigate noisy prediction and it works better on datasets with more semantic categories. This is not a surprise because more categories cause heavier prediction noise and hence the LD works more effectively. Table \ref{Table:AblationStudy} shows the significant performance gain (7 percent on COCO-Stuff and almost 10 percent on PASCAL-Context) from the baseline by applying the proposed SVC.

To further study where the performance gain of the proposed SVC comes from, we compare it with the shape-fixed context (SFC) that is implemented by setting a constant shape mask, i.e., $\mathcal{M}_{m,n}^{i,j}$$=$$1$ in Eq.$\!$ (\ref{eq3}). We compare them on PASCAL-Context with different kernel sizes shown in Table \ref{Table:LargeKernelSize}, where the 0$\times$0 means the baseline without the context layer.  As we employ just a single layer to capture the context information, all kernels used in Table \ref{Table:LargeKernelSize} are larger than convolutional kernels of most other work so that diverse shapes and scales of contextual information could be modelled in a single layer. Table \ref{Table:LargeKernelSize} shows that with the increase of kernel size, the segmentation performance improves up to a certain level then slightly drops with further increase of the kernel size. This is because the network lose too much locality information in overlarge kernel situation. It also shows that a simple increase of the network parameters may not always improve the performance. Table \ref{Table:LargeKernelSize} shows that the proposed SVC significantly outperforms the SFC at all different kernel sizes. It is also not a surprise that the best performance of the SVC is achieved at the kernel size ($23$$\times$$23$), much larger than that of the SFC ($15$$\times$$15$) because the proposed SVC provides diverse shape context, which is implicitly multi-scale with better location identity.

\subsection{Visualization of the semantic shape mask}

\begin{figure}[t]
\centering
  \includegraphics[width=0.47\textwidth]{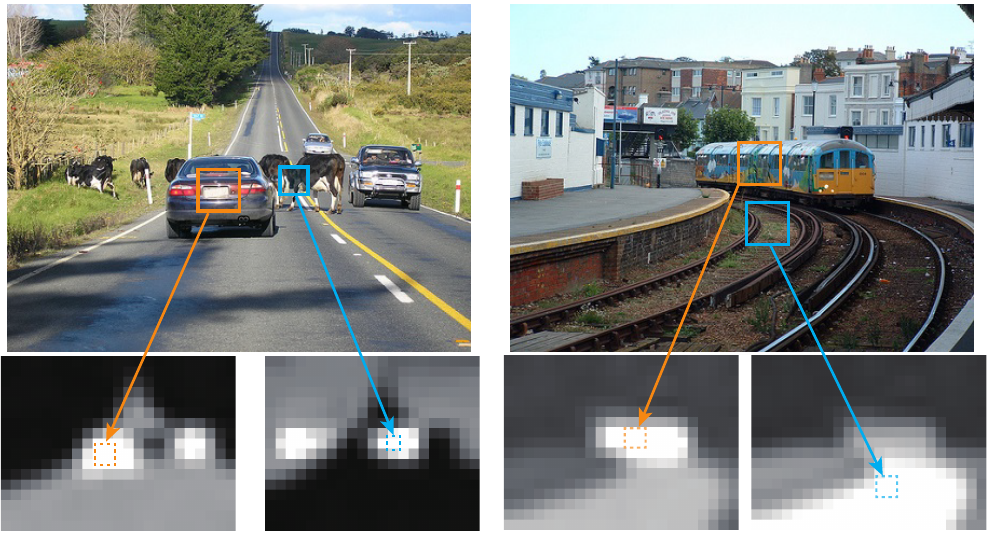}
\vspace{-0.1cm}
\caption{Four visual examples of the shape-variant masks $\mathcal{M}_{m,n}^{i,j}$ generated at four different locations of two testing images by the learned network. The mask center ($i,j$) is indicated by the small square and its value at ($m,n$) within the image is shown by the gray level.}
\label{Figure:shapemasks_visual}
\vspace{-0.2cm}
\end{figure}

As Table \ref{Table:AblationStudy} and Table \ref{Table:LargeKernelSize} show the significant performance gains brought by the proposed shape-variant context (SVC) that is determined by the proposed shape mask $\mathcal{M}_{m,n}^{i,j}$, it is worthwhile to further study how the mask captures the shape of the context by visualizing it. Four examples of the shape mask $\mathcal{M}_{m,n}^{i,j}$ generated at four different locations of two testing images by the learned network are shown in Figure \ref{Figure:shapemasks_visual}. The first is the mask of the center of a car. It has higher values at pixels of cars and road as they contains the context information of the center of the car. The second is the mask of some pixel of the cow in the middle of a road. It has higher values at pixels of cows and grass, though they are far away from the target cow and separated by the road. Values of the second mask are low in the area of road as it does not show correlation with the cow in the training database. Consistently, the semantic correlation of the context is also shown by the third and fourth masks respectively for a train and a railway in the second testing image.

\subsection{Comparison with the State-of-the-Arts}
\label{Section:BenchMarkResults}
\begin{figure}[t]
\centering
  \scriptsize{\hspace{2.5em} \textbf{Images}  \hfill \hspace{1em} \textbf{Baseline} \hfill \hspace{-0.2em} \textbf{SVCNet (ours)} \hfill \hspace{-1.76em} \textbf{Ground Truth} \hspace{0.5em} } \\
 \includegraphics[width=0.476\textwidth]{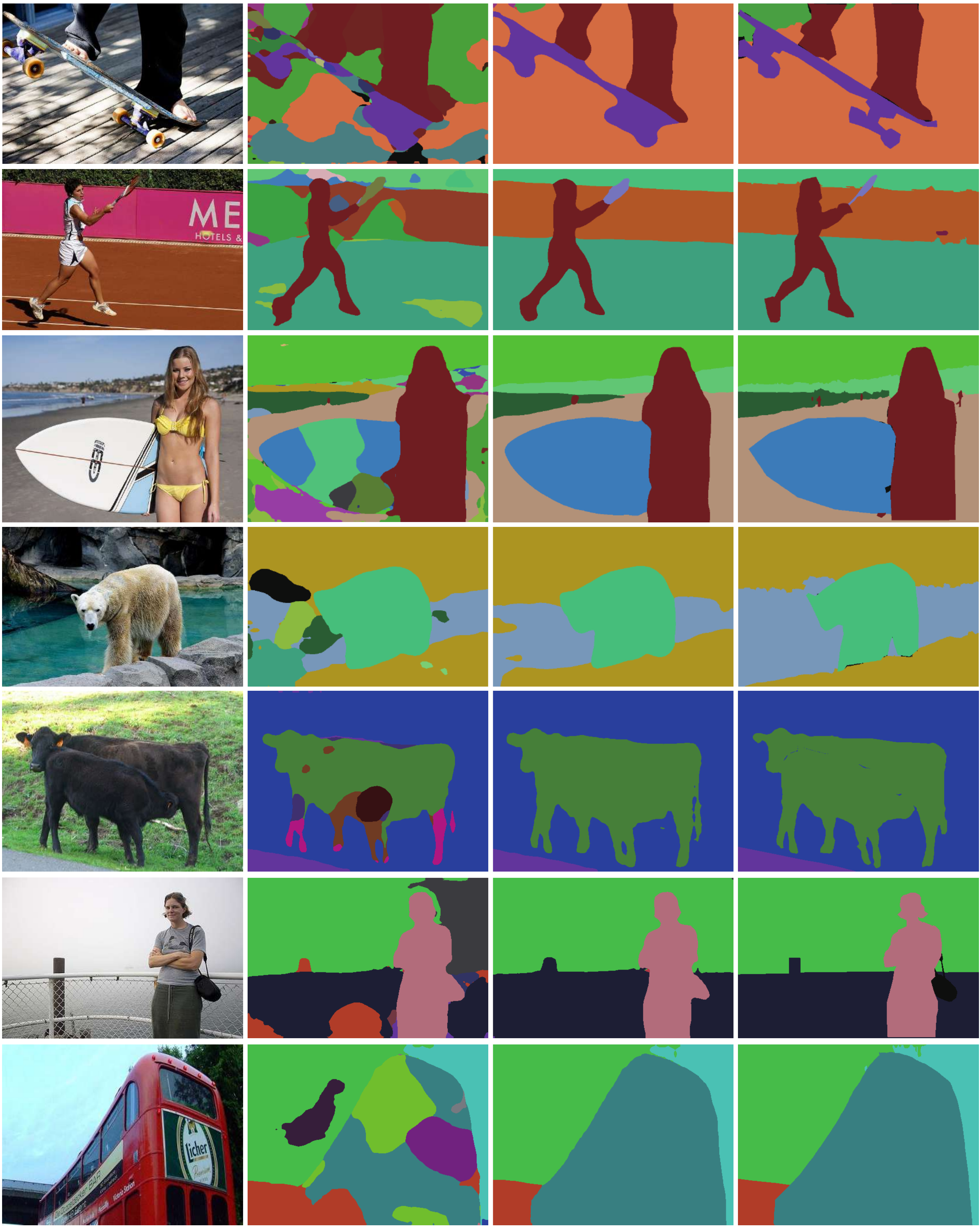}
\caption{Qualitative segmentation examples on COCO-Stuff (1st-4th rows) and PASCAL-Context (5th-7th rows).}
\vspace{-0.24cm}
\label{Figure:QualitativeResults}
\end{figure}

The proposed semantic segmentation network is named as SVCNet and we compare it with the state-of-the-arts on six public benchmark, COCO-Stuff, SIFT-Flow, CamVid, PASCAL-Person-Part, PASCAL-Context, and Cityscapes. Before the quantitative comparison, some qualitative results of the proposed SVCNet are shown in Figure \ref{Figure:QualitativeResults}.

\vspace{0.24em}
\noindent\textbf{COCO-Stuff}~\cite{Caesar_2018_CVPR} provides dense pixel-wise annotations for 171 semantic categories. There are 9000 images used for training and 1000 images used for testing. Quantitative results of COCO-Stuff are shown in Table \ref{Table:COCOStuff_results_comparison}. The proposed SVCNet outperforms the previous state-of-the-arts across all evaluation metrics.

\begin{table}[t]
\footnotesize
\begin{center}
\begin{tabular}{lcccc}
\toprule
Methods & pixel acc. & mean acc. & mean IoU \\
\midrule
FCN~\cite{Caesar_2018_CVPR} & 52.0 & 34.0 & 22.7 \\
DeepLab~\cite{chen2015semantic} & 57.8 & 38.1 & 26.9 \\
FCN-8s~\cite{long2015fully} & 60.4 & 38.5 & 27.2 \\
DAG-RNN+CRF~\cite{shuai2017scene} & 63.0 & 42.8 & 31.2 \\
DC+FCN+~\cite{hu2017labelbank} & 65.5 & 44.6 & 33.6 \\
Deeplab-V2~\cite{chen2016deeplab} & 65.1 & 45.5 & 34.4 \\
CCL-ResNet101~\cite{Ding_2018_CVPR} & 66.3 & 48.8 & 35.7 \\
DSSPN~\cite{liang2018dynamic} & 68.5 & 48.1 & 36.2 \\
\midrule
SVCNet (ours) & \textbf{69.2} & \textbf{51.5} & \textbf{39.6} \\
\bottomrule
\end{tabular}
\end{center}
\vspace{-0.2cm}
\caption{COCO-Stuff testing accuracies.}
\label{Table:COCOStuff_results_comparison}
\end{table}

\vspace{0.24em}
\noindent\textbf{SIFT-Flow}~\cite{liu2009nonparametric} contains 2688 images annotated with 33 semantic classes. There are 2488 training images and 200 testing images. Quantitative results are shown in Table \ref{Table:SiftFlow_results_comparison}. The proposed SVCNet outperforms previous state-of-the-arts across all evaluation metrics.
\begin{table}[t]
\footnotesize
\begin{center}
\begin{tabular}{lccc}
\toprule
Methods & pixel acc. & mean acc. & mean IoU \\
\midrule
Liu et al.~\cite{liu2011sift} & 76.7 & - & - \\
Tighe et al.~\cite{tighe2013finding} & 75.6 & 41.1 & - \\
Farabet et al.~\cite{farabet2013learning} & 78.5 & 29.6 & - \\
Pinheiro et al.~\cite{pinheiro2014recurrent} & 77.7 & 29.8 & - \\
Sharma et al.~\cite{sharma2014recursive} & 79.6 & 33.6 & - \\
Yang et al.~\cite{yang2014context} & 79.8 & 48.7 & - \\
FCN-8s~\cite{shelhamer2016fully} & 85.9 & 53.9 & 41.2 \\
DAG-RNN+CRF~\cite{shuai2017scene} & 87.8 & 57.8 & 44.8 \\
Piecewise~\cite{lin2016efficient} & 88.1 & 53.4 & 44.9 \\
\midrule
SVCNet (ours) & \textbf{89.1} & \textbf{58.2} & \textbf{46.3} \\
\bottomrule
\end{tabular}
\end{center}
\vspace{-0.2cm}
\caption{SIFT-Flow testing accuracies.}
\label{Table:SiftFlow_results_comparison}
\end{table}

\begin{table}
	\begin{minipage}[b]{0.5\linewidth}
		\centering
        \footnotesize
        \begin{center}
        \begin{tabular}{lc}
        \toprule
        Methods &  mean IoU \\
        \midrule
        DeconvNet~\cite{noh2015learning} &48.9  \\
        SegNet~\cite{badrinarayanan2017segnet}  &50.2 \\
        DeepLab~\cite{chen2015semantic}  &54.7 \\
        DilatedNet~\cite{yu2015multi} &65.3 \\
        Dilation+FSO~\cite{kundu2016feature}&66.1 \\
        FC-DenseNet~\cite{jegou2017one} &66.9 \\
        G-FRNet~\cite{islam2017gated} &68.0\\
        DenseDecoder~\cite{Bilinski_2018_CVPR} &70.9\\
        \midrule
        SVCNet (ours) &\textbf{75.4}  \\
        \bottomrule
        \end{tabular}
        \end{center}
        \vspace{-0.2cm}
        \caption{CamVid.}
        \vspace{-0.24cm}
        \label{Table:resultsCamVid}
	\end{minipage}
	\begin{minipage}[b]{0.48\linewidth}
      \footnotesize
      \begin{center}
      \begin{tabular}{lc}
      \toprule
      Methods &  mean IoU \\
      \midrule
      Attention~\cite{chen2016attention} &56.4  \\
      HAZN~\cite{xia2016zoom}  &57.5 \\
      LG-LSTM~\cite{liang2016semantic}  &58.0 \\
      Graph LSTM~\cite{liang2016ECCV}  &60.2 \\
      DeepLab~\cite{chen2015semantic}  &62.8 \\
      DeepLab-V2~\cite{chen2016deeplab} &64.9 \\
      RefineNet~\cite{Lin_2017_CVPR} &68.6\\
      DenseDecoder~\cite{Bilinski_2018_CVPR} &68.6\\
      \midrule
      SVCNet (ours) &\textbf{73.9}  \\
      \bottomrule
      \end{tabular}
      \end{center}
      \vspace{-0.2cm}
      \centerline{\caption{\footnotesize{PASCAL-Person-Part}.}\label{Table:resultsofPASCAL-Person-Part}}

      \vspace{-0.24cm}
	\end{minipage}
\end{table}

\vspace{0.24em}
\noindent\textbf{CamVid}~\cite{brostow2008segmentation} is a road scene image segmentation dataset which provides dense pixel-wise annotations for 11 semantic categories.  There are 367 training images, 101 validation images and 233 testing images. The testing results are shown in Table \ref{Table:resultsCamVid}. It shows that the proposed SVCNet outperforms previous state-of-the-arts by a large margin.

\vspace{0.24em}
\noindent\textbf{PASCAL-Person-Part}~\cite{chen_cvpr14} provides pixel-level labels for six person parts. There are 1717 training/validation images and 1818 testing images. Quantitative results of PASCAL-Person-Part are reported in Table~\ref{Table:resultsofPASCAL-Person-Part}. It shows that the proposed SVCNet outperforms the previous state-of-the-arts by a large margin on this small dataset, which indicates that the the proposed approach could be trained very well even on small dataset.


\vspace{0.24em}
\noindent\textbf{PASCAL-Context}~\cite{mottaghi_cvpr14} provides pixel-wise segmentation annotation for 59 classes. There are 4998 training images and 5105 testing images. Quantitative results of Pascal Context are shown in Table \ref{Table:resultsContext}. It shows that the proposed SVCNet outperforms the state-of-the-arts by a large margin.

\begin{table}[t]
\footnotesize
\begin{center}
\begin{tabular}{p{5.76cm}p{1.26cm}<{\centering}}
\toprule
Methods & mean IoU \\
\midrule
FCN-8s~\cite{shelhamer2016fully}  & 39.1 \\
CRF-RNN~\cite{zheng2015conditional} & 39.3 \\
BoxSup~\cite{dai2015boxsup} & 40.5 \\
HO-CRF~\cite{arnab2016higher}& 41.3 \\
Piecewise~\cite{lin2016efficient} & 43.3 \\
FCRN~\cite{wu2016bridging} & 44.5 \\
EFCN~\cite{shuai2019toward}& 45.0 \\
DeepLab-V2\cite{chen2016deeplab} & 45.7 \\
Global-Context\cite{hung2017scene}  & 46.5 \\
RefineNet-ResNet152~\cite{Lin_2017_CVPR} & 47.3 \\
DenseDecoder~\cite{Bilinski_2018_CVPR} & 47.8 \\
MSCI~\cite{Lin_2018_ECCV} & 50.3 \\
CCL-ResNet101~\cite{Ding_2018_CVPR}& 51.6 \\
EncNet~\cite{Zhang_2018_CVPR}& 51.7 \\
\midrule
SVCNet (ours)& \textbf{53.2} \\
\bottomrule
\end{tabular}
\end{center}
\vspace{-0.12cm}
\caption{PASCAL-Context testing accuracies.}
\label{Table:resultsContext}
\end{table}

\vspace{0.24em}
\noindent\textbf{Cityscapes}~\cite{cordts2016cityscapes} contains 5000 street scene images with pixel-level fine annotations and 19 classes are considered for evaluation. There are 2975 training images, 500 validation images and 1525 testing images. The test results are shown in Table \ref{Table:Cityscapes_results_comparison}.
\begin{table}[t]
\footnotesize
\begin{center}
\begin{tabular}{p{5.76cm}p{1.26cm}<{\centering}}
\toprule
Methods & mean IoU \\
\midrule
Deeplab-v2~\cite{chen2016deeplab} & 70.4 \\
RefineNet-Res101~\cite{Lin_2017_CVPR} &73.6 \\
DSSPN-Universal~\cite{liang2018dynamic} &76.6 \\
GCN~\cite{Peng_2017_CVPR} & 76.9 \\
DepthSet~\cite{kong2018recurrent}& 78.2 \\
PSPNet~\cite{Zhao_2017_CVPR}& 78.4 \\
AAF~\cite{ke2018adaptive}& 79.1 \\
DFN~\cite{Yu_2018_CVPR}& 79.3 \\
PSANet~\cite{Zhao_2018_ECCV}& 80.1 \\
DenseASPP-DenseNet161~\cite{yang2018denseaspp}&80.6 \\
\midrule
SVCNet (ours)& \textbf{81.0} \\
\bottomrule
\end{tabular}
\end{center}
\vspace{-0.12cm}
\caption{Cityscapes testing accuracies.}
\vspace{-0.24cm}
\label{Table:Cityscapes_results_comparison}
\end{table}

\section{Conclusion}
\vspace{-0.1cm}
\begin{spacing}{0.96}
In this work, we propose to aggregate the context information based on the semantic correlation rather than the predefined spatial-dependent window to collect more effective and discriminative surrounding information for semantic segmentation. The semantic-correlated information even at a far away spatial location will be enhanced and the semantic-uncorrelated information even at a close spatial location will be suppressed in collecting the context information. To this end, we first propose a novel paired convolution to learn the feature semantic-correlation from the training images and to infer it of the query image. This generates a semantic shape mask at each position of the image. Based on it, we propose a shape-variant convolution, in which the receptive field of the convolution is specified by different semantic shape masks at different positions of different query images. The semantic shape masks form diverse scales and shapes of the convolution receptive field to aggregate discriminative context information effectively. Furthermore, to ease the labeling errors, we propose a labeling denoising model, which utilizes more robust higher-level features to attenuate the prediction errors caused by noisier lower-level features. Without bells and whistles, the proposed segmentation network achieves new state-of-the-arts consistently on the six public semantic segmentation datasets, COCO-Stuff, SIFT-Flow, CamVid, PASCAL-Person-Part, PASCAL-Context, and Cityscapes.
\end{spacing}
\vspace{-0.24cm}
\section*{Acknowledgement}
\vspace{-0.2cm}
\begin{spacing}{0.89}
\footnotesize{This research is supported by Singapore Ministry of Education Academic Research Fund Grant no: 2015-T1-002-140, MOE Tier 1 RG 123/15. It is also supported by the BeingTogether Centre, a collaboration between Nanyang Technological University (NTU) Singapore and University of North Carolina (UNC) at Chapel Hill. The BeingTogether Centre is supported by the National Research Foundation, Prime Minister's Office, Singapore under its International Research Centres in Singapore Funding Initiative. We gratefully acknowledge the support of NVIDIA Corporation with the donation of the Titan Xp GPU used for this research.}
\end{spacing}

\begin{spacing}{0.951}
{\small
\bibliographystyle{ieee_fullname}
\bibliography{cvpr19}
}
\end{spacing}
\end{document}